%% file: main.tex
\documentclass[10pt,twocolumn,letterpaper]{article}

\usepackage{cvpr}
\usepackage{times}
\usepackage{epsfig}
\usepackage{graphicx}
\usepackage{amsmath}
\usepackage{amssymb}
\usepackage{algorithm}
\usepackage{algorithmic}

\usepackage[pagebackref=true,breaklinks=true,letterpaper=true,colorlinks,bookmarks=false]{hyperref}

 \cvprfinalcopy 

\def\cvprPaperID{****} 

\ifcvprfinal\fi
\begin{document}

\title{ On Image segmentation using Fractional Gradients-\\Learning Model Parameters using  Approximate Marginal Inference}

\author{Anish Acharya\\
Schlumberger \\
{\tt\small AAcharya2@slb.com}
\and
Dr. Uddipan Mukherjee\\
Intel Corporation\\
{\tt\small umukherjee@uci.edu}
\and
Dr. Charless Fowlkes\\
University of California Irvine\\
{\tt\small fowlkes@uci.edu}
}
\maketitle
\begin{abstract}
Estimates of image gradients play a ubiquitous role in image segmentation and classification problems since gradients directly relate to the boundaries or the edges of a scene. This paper proposes an unified approach to gradient estimation based on fractional calculus that is computationally cheap and readily applicable to any existing algorithm that relies on image gradients.  We show experiments on edge detection and image segmentation on the Stanford Backgrounds Dataset where 
these improved local gradients outperforms state of the art, achieving a performance of 79.2\% average accuracy.

\end{abstract}

\section*{Keywords}
Image Segmentation, Fractional Derivative

\input{introduction}

\input{imsegment}

\input{learning}

\input{results}

{\small
\bibliographystyle{ieee}
\bibliography{mybib}
}

\end{document}

%% file: introduction.tex
\section{Introduction}
\label{sec:introduction}

A huge variety of modern computer vision algorithms are built on the computation of local derivatives of image brightness which provide local discriminative cues to scene structure, e.g., texture based segmentation~\cite{Malik:2001:CTA:543015.543016}, HOG based~\cite{1467360}, and SIFT~\cite{Lowe:2004:DIF:993451.996342} frameworks. Typically an input image is convolved with an image filter bank consisting of gradient filters in different orientations and scales. The response of these filters are used to construct a feature space which embeds structural image information.
This approach is found to be incredibly powerful for detecting key image features or recognizing objects over different scales, appearances and poses. The success of these algorithms prove the importance of gradient orientations as a robust feature for  recognition~\cite{Shi:2000:NCI:351581.351611,Malik:1999:TCR:850924.851546, Malik90preattentivetexture, Blake:1987:VR:30394}. 
To date the most widely used gradient based features are extracted using the First Order Derivative of a Gaussian (DOG) kernel taken in different orientation and scales or considering the zero crossing in the response to the Laplacian of Gaussian Operator (LOG). As the order of the gradient operator increases, the response to edges is also higher, but an increased order also gives an undesirably higher response to noise or isolated points. In order to avoid noise enhancement images are often smoothed prior to edge detection. However, smoothing can cause a loss of structural information in some cases. Thus, choosing the proper gradient order for a given image depends largely on the image, e.g. we might want a weaker edge response for a noisy image and a stonger one for a crisp image.


\begin{figure}[h]
\centering
\includegraphics[scale=0.37]{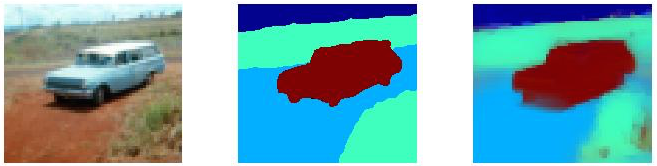}
\caption{\label{fig:results1}Image segmentation using FDOG edge detector. From left to right: original image, true label and estimated label.}
\end{figure}

\begin{figure*}[ht]
\centering
\includegraphics[width=3.2cm,height=4.4cm]{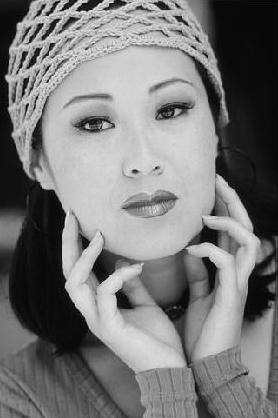}
\includegraphics[width=3.2cm,height=4.4cm]{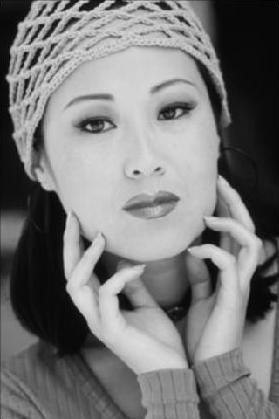}
\includegraphics[width=3.2cm,height=4.5cm]{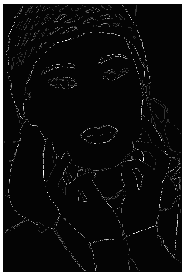}
\includegraphics[width=3.2cm,height=4.5cm]{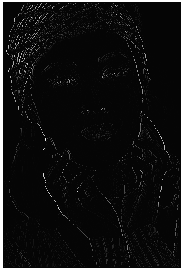}
\includegraphics[width=3.2cm,height=4.5cm]{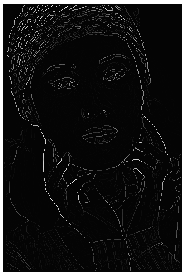}
\caption{\label{fig:quickcomp}Edge detection using FDOG: Left to right: sample image, smoothed image, manual edge detection, DOG edge detection and FDOG with v=0.5. It can be observed that FDOG performance is closer to manual detection than DOG. }
\end{figure*}

In order to address this problem we propose a framework for generating a family of gradient operators with fractional orders, and show that the fractional gradients work better than integral ones for edge detection and eventual image segmentation. In order to design the fractional gradient operators, we use concepts from Fractional Calculus, which has been widely and efficiently used in the fields of signal processing. The fractional orders introduce higher order hyper-damped and ultra-damped poles which lends much more flexibility and accuracy in modeling complicated system dynamics~\cite{Acharya:2014:ECA:2537167.2537260,XX}.

In essence, we use fractional order gradients for designing fractional order derivative of Gaussian (FDOG) filters and show that our design achieves better edge detection performance than existing integer order gradient filters (Figure~\ref{fig:quickcomp}). In order to validate this performance gain, we apply our method to image segmentation (on images from Stanford Backgrounds Dataset~\cite{Koller:2009:PGM:1795555}, an example shown in Figure~\ref{fig:results1}) by building Histogram of Gradient Orientation (HOG) features on top of the FDOG filters, and learning a Markov Random Field based Model using Tree Re-weighted Belief Propagation based on the approach proposed in~\cite{10.1109/TPAMI.2013.31}. 

The rest of the paper is organised as follows: Section~\ref{sec:imsegment} gives a detailed description of the design of FDOG filters and demonstartes its performance on test images. Section~\ref{sec:learning} describes how the FDOG based approach can be used for image segmentation and Section~\ref{sec:results} depicts the performance gain for the same over existing methods, followed by a conclusion in Section~\ref{sec:conclusion}.

%% file: imsegment.tex
\section{Fractional Order Derivative of Gaussian Filters for Edge Detection}
\label{sec:imsegment}

The most fundamental structural information embedded in an image is its edges. Edge detection is a well established notion in Computer Vision and is still a significant area of research. Among the various algorithms used for edge detection, detectors like Canny~\cite{Canny:1986:CAE:11274.11275}, Sobel, Pretwitt, Robert's Cross are particularly popular, Canny being the most reliable and widely used in the community.
Recently, the works of Perona, Malik et.al., based on  Anisotropic Diffusion algorithm~\cite{Perona:1990:SED:78302.78304}, and a new edge detection scheme proposed by Harris et.al.~\cite{Harris88alvey},  also gained popularity. However, one major disadvantage of Harris Detector is that it is not scale invariant. Though there has been many recent advances on the improvement of this~\cite{Brown:2005:MMU:1068507.1068957,Chum:2005:MPP:1068507.1068918}, most of these techniques are computationally complex compared to the Canny as well as Anisotropic Diffusion based detectors. These detectors might be extremely efficient when the sole task is edge detection. However, when we are concerned with a broader problem such as segmentation, edge detection is one of the many steps involved, and we need a robust and efficient detector without much overhead of computational complexity. 
An optimal edge detector captures the underlying geometric structure of the image (i.e. detects all the edges successfully) without being much affected by the noise present. As has been already shown by Canny et.al.~\cite{Canny:1986:CAE:11274.11275}, it is impossible to realize such an optimal filter for edge detection, although the first order gradient of Gaussians is a good approximation with around  20\% error. The order of the filter gradient is directly proportional to the number of detected edges and the amount of detected noise. Thus we get more peaked response at an edge with the second order LOG (Laplacian of Gaussian) as compared to the first order DOG (derivative of gaussian), with an additional side-effect of a higher response to noisy peaks for LOG as compared to DOG. In order to control this trade-off more efficiently, we propose a detector with fractional order gradient operator (FDOG) and show that the edge detection performance can be improved by using FDOG. 


The FDOG filters are designed using concepts from fractional order calculus. Similar filters have been designed in the frequency domain and has been practiced frequently in various control applications. Acharya et.al.~\cite{Acharya:2014:ECA:2537167.2537260} proposed a structured exact implementation of arbitrarily chosen cut-off based filters. We extend this idea to design a 2-D convolution mask based on the general discrete approximation of the n-th order differentiation operator, represented by (~\ref{eq:eq1}).

\begin{figure*}[ht]
\centering
\includegraphics[width=4.1cm,height=4.6cm]{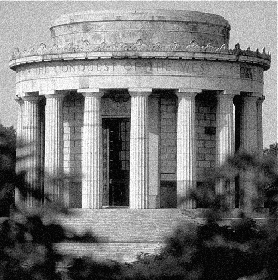}
\includegraphics[width=4.1cm,height=4.6cm]{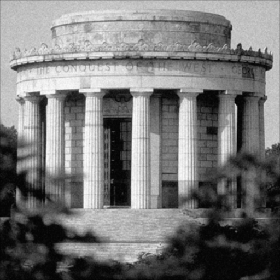}
\includegraphics[width=4.1cm,height=4.6cm]{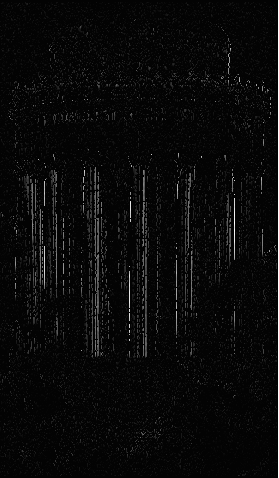}
\includegraphics[width=4.1cm,height=4.6cm]{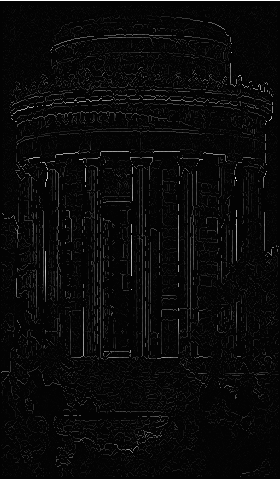}

\caption{\label{fig:quickcomp2}Edge detection using FDOG on a noisy image. From left to right: original image, smoothed with $\sigma=2$ Gaussian kernel, DOG edge detection and FDOG edge detection. It can be observed that FDOG outperforms DOG even on a noisy image.}
\end{figure*}

\begin{multline}
\label{eq:eq1}
\frac{\partial^vs(x,y)}{\partial x^v}=s(x,y)-vs(x-1,y) + 
\frac{v(v-1)}{2}s(x-2,y) \\-\frac{v(v-1)(v-2)}{6}s(x-3,y)+... ,\\
\frac{\partial^vs(x,y)}{\partial y^v}=s(x,y)-vs(x,y-1) + 
\frac{v(v-1)}{2}s(x,y-2) \\-\frac{v(v-1)(v-2)}{6}s(x,y-3)+...
\end{multline}

The gradient to be computed is thus given by the sum of the partial derivatives (Equation~\ref{eq:eq2}).
\begin{equation}
\label{eq:eq2}
\bigtriangledown^v=\frac{\partial^vs(x,y)}{\partial x^v}+\frac{\partial^vs(x,y)}{\partial y^v}=\bigtriangledown^v_x+\bigtriangledown^v_y
\end{equation}
where instead of restricting the order to be integer i.e. in case of DOG, $v=1$ and LOG $v=2$; we propose to let $v$ take any fractional value. Theoretically, $v$ should be limited to one decimal place to avoid mathematical intractability. Also, for efficiency, we restrict the number of terms in equation~\ref{eq:eq1} to three. Figure~\ref{fig:quickcomp} shows the result of applying FDOG of order $0.5$ for edge detection on a standard image taken from the Berkeley Segmentation Dataset. Figure~\ref{fig:quickcomp} also shows a DOG kernel based edge detection and a benchmark obtained by averaging over the five hand edge-detected versions of the same image. All the results are subjected to non maximal suppression and displayed in normalized scale for comparison purposes. As can be observed, FDOG of order $0.5$ seems to perform better than standard DOG operator visually. In other words, FDOG weeds out noise and reveals structural image information more clearly. Even in highly noisy images, FDOG seems to perform better, as is verified on an image (taken from the CSIQ image database \cite{larson2010most}) corrupted with white noise (Figure~\ref{fig:quickcomp2}).

\subsection{Evaluating FDOG Edge detection}
The standard measure to evaluate the performance of an edge detector is to measure the Peak Signal-to-noise Ratio (PSNR)~\cite{Canny:1986:CAE:11274.11275}. This is an effective measure when the noise models of an image are provided, or more generally if the imaging circumstances are known before hand and the detected edges can be clearly classified as true edges or noise. A more practical and widely adopted approach for any arbitrary given image is to filter the image and measure the structural loss caused therby. It assumes a general framework of the corrupted image as given in equation~\ref{eq:snr2}. 

\begin{equation}
\label{eq:snr2}
x[n]=s[n]+w[n]
\end{equation}

where $x[n]$ is the given image and $s[n]$ is the filtered one. The PSNR is then defined as shown in Equation~\ref{eq:snr3}.
\begin{equation}
\label{eq:snr3}
\begin{split}
PSNR=10log_{10}\frac{max(s^2[n])}{MSE}, 
\\
MSE=\frac{1}{MN}\sum_{i=1}^{M}\sum_{j=1}^{N}[\hat{s}(i,j)-s(i,j)]^{2}
\end{split}
\end{equation}

where $s[n]$ is the maximum signal value, typically $255$ when considering image pixel values, $M$ and $N$ are the height and width of the image, $\hat{s}$ is the processed image and $s$ is the original image, and MSE is the mean squared error.
While PSNR gives us an idea of how much structural information is retained in the gradient operated image, another criterion of good detection is to minimize the effect of noise. This can be addressed using the idea of False Discovery rate (DE)~\cite{citeulike:1042553}, which is given in equation~\ref{eq:DE}.
\begin{equation}
\label{eq:DE}
DE=P_m*P_f 
\end{equation}
where $P_m$ and $P_f$ are respectively Probability of missed detections and probability of false positives. A low value of this product implies good detection i.e. better edge detector.   

\subsubsection{Finding the optimal Gradient order}
The correct order of the gradient to be used in FDOG is data specific and should be optimized over all the images in the dataset. Thus finding the optimal order of the derivative can be now viewed as an optimization task with th score function, $J=\frac{PSNR}{DE}$ to be maximized.
Any standard approach like gradient descent can be used to find the optimal order. We restrict the search space to within one decimal place i.e. $v$ can only take values in steps of $0.1$ and select the order with highest score.
Applying our algorithm to various datasets like the berkley segmentation dataset~\cite{937655}, The Stanford backgrounds dataset~\cite{5459211} and CSIQ noisy database it is found that in most cases the optimal order turns out be in the range 0.5-0.8 with lower order for highly noisy images like astronomical images. 
\begin{figure}[h]
\centering
\includegraphics[scale=0.32]{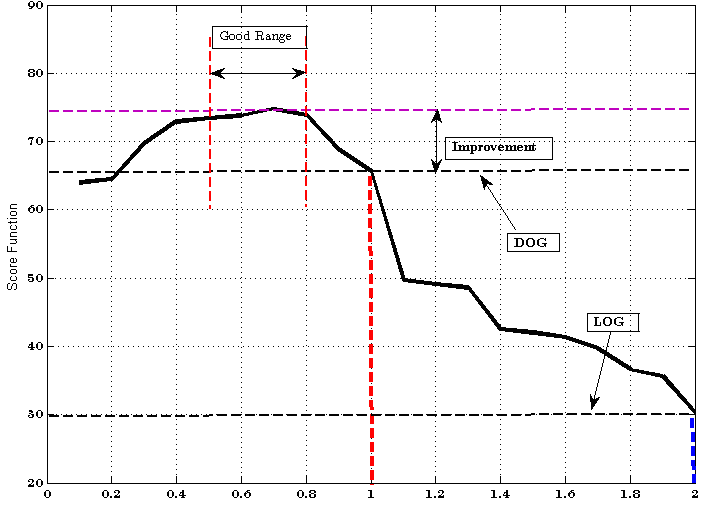}
\caption{\label{fig:optimalorder}Performance of different Gradient Orders on Stanford Backgrounds Dataset}
\end{figure}

Figure~\ref{fig:optimalorder} shows the plot of score function with respect to different filter orders applied on the Stanford background dataset. The score function for each order have been computed over all the images and averaged to produce the reported value. It can be seen that the approximate optimal order is somewhere around 0.5-0.8.



\begin{figure}
\centering
\includegraphics[scale=0.235]{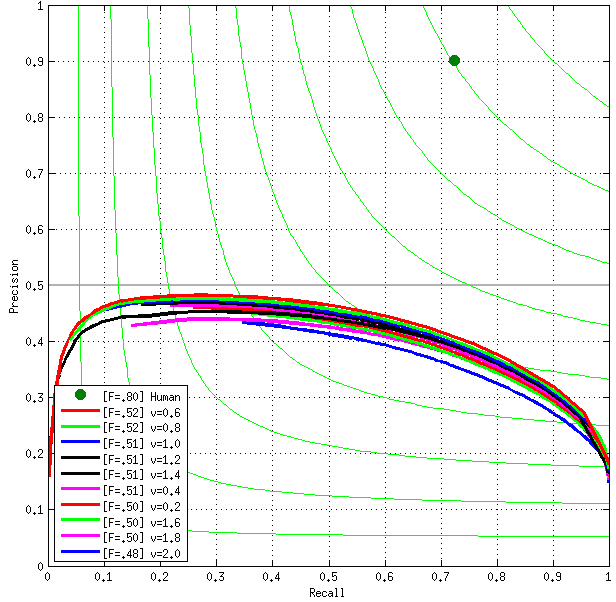}
\includegraphics[scale=0.235]{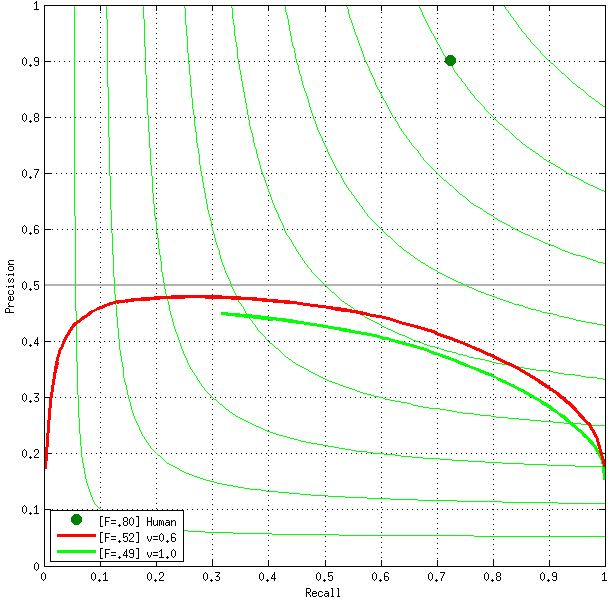}
\caption{\label{fig:precrec}Top:left: Precision Recall curves on training BSDS500 for Varying Gradient Order right: comparing PR curves for FDOG order 0.6 and DOG}
\end{figure}

In order to validate our scoring scheme, we also generate a precision-recall curve for detection (Figure~\ref{fig:precrec}, Table~\ref{table:table1}). First we threshold the generated boundary map into several levels and at each level compute the precision, which is the probability that a machine generated pixel is a boundary pixel, and recall, which is the probability that a true boundary pixel is detected~\cite{937655}. The precision-recall curve reflects the inherent trade-off between misses and false positives. It would be nice to be able to compare the curves for different algorithms in terms of a single number. In an attempt to do so we introduce the F-measure, which is the harmonic mean of precision and recall, as a yardstick for an algorithm's performance. In general, for two algorithms whose curves are roughly parallel, the one furthest from the origin dominates the other. The F-measure is defined at all points on the precision-recall curve.  We report the maximum F-measure value across an algorithm's precision-recall curve as its summary statistic.
These results should not be considered as edge detection results since these are based on just computing the raw local gradients with different orders using \ref{eq:eq1} and does not apply any processing before feeding to the benchmarking and thus the results would be lower than that in the BSDS leader-board. However, we use these PR curves as a way of comparing the descriptive power of different gradient orders to validate our formulated evaluation metric based on PSNR and FDR.

Figure~\ref{fig:precrec} shows the precision-recall curves for varying orders and the variation of F-number with varying filter orders.  Intuitively, and also as the curves show, an increased order enhances more noisy peaks and too small an order fails to mine structural information. From the training images it is observed that 0.6 gives the best F-score of 0.523 which is almost 1\% improve over DOG. Whereas the LOG performs way bad and gives an F-score of 0.480 - almost 4\% less than the best. We use the Filter order as 0.6 on the test images to generate the precision-recall curve and also provide the precision-recall of the standard DOG for comparison purposes in Figure~\ref{fig:precrec}.


\begin{table}[h]
\centering
\begin{tabular}{ |c|c|c|c| } 
 \hline
Gradient Order &  ODS & OIS & AP \\ 
 \hline
0.2 & 0.501 & 0.535 & 0.288 \\ 
0.4 & 0.506 & 0.538 & 0.311 \\ 
0.6 &  0.523 & 0.555 & 0.311 \\ 
0.8 &   0.518 & 0.552 & 0.404 \\ 
1 &    0.514 & 0.549& 0.370 \\ 
1.2 &  0.510 & 0.544 & 0.338 \\ 
1.4 & 0.508 & 0.546 & 0.266 \\ 
1.6 & 0.497 & 0.530 & 0.419 \\ 
1.8 & 0.495 & 0.536 & 0.329 \\ 
2 & 0.480 & 0.513 & 0.232 \\ 
  
 \hline
\end{tabular}
\caption{\label{table:table1} Precision-recall and F-measures}
\end{table}

\begin{figure}[h]
\centering
\includegraphics[height=3cm,width=4.1cm]{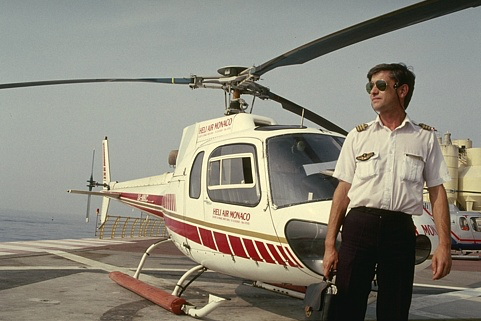}
\includegraphics[height=3cm,width=4.1cm]{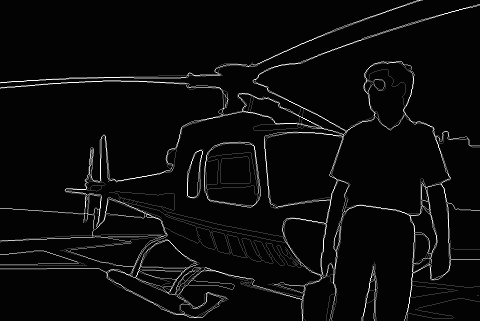}
\includegraphics[height=3cm,width=4.1cm]{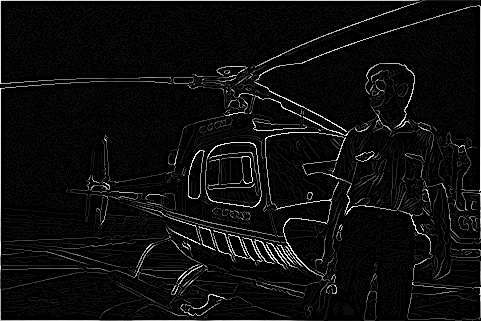}
\includegraphics[height=3cm,width=4.1cm]{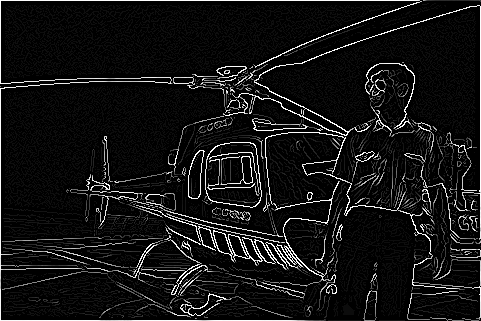}
\includegraphics[height=3cm,width=4.1cm]{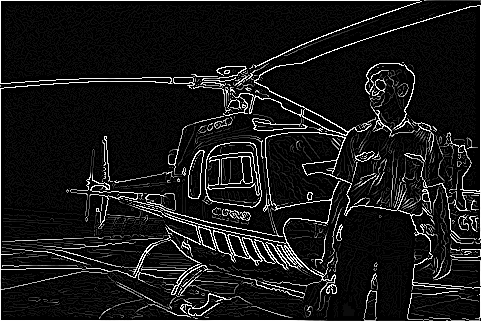}
\includegraphics[height=3cm,width=4.1cm]{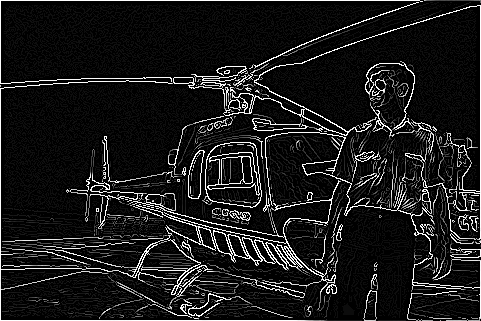}
\includegraphics[height=3cm,width=4.1cm]{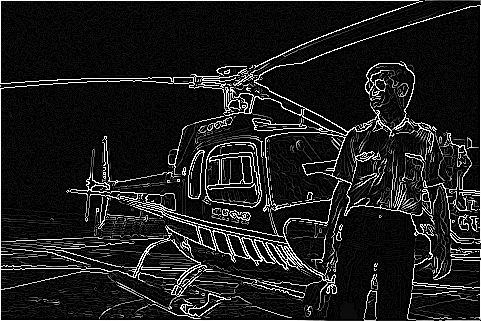}
\includegraphics[height=3cm,width=4.1cm]{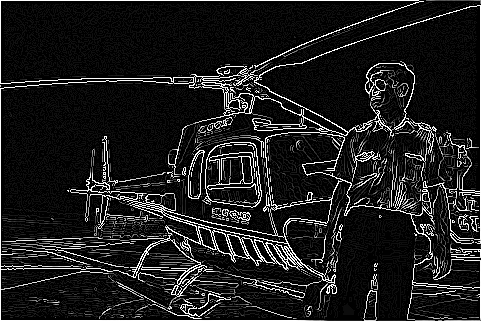}
\includegraphics[height=3cm,width=4.1cm]{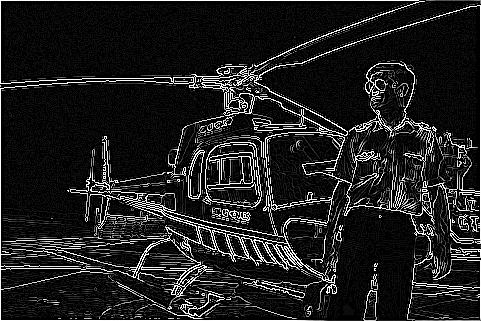}
\includegraphics[height=3cm,width=4.1cm]{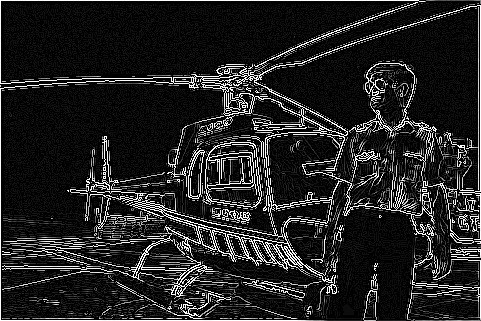}
\includegraphics[height=3cm,width=4.1cm]{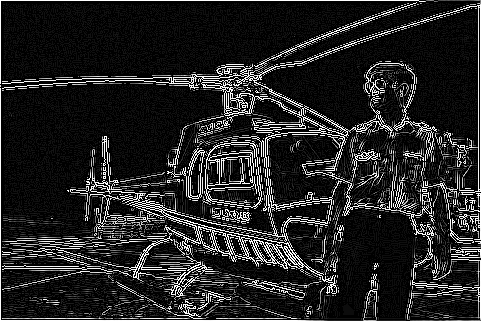}
\includegraphics[height=3cm,width=4.1cm]{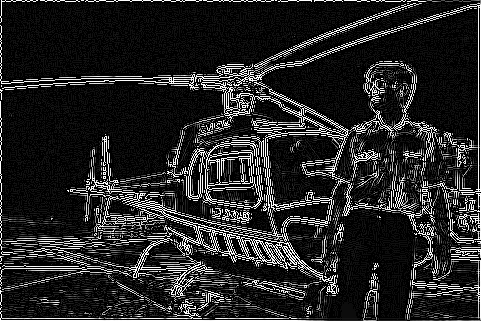}
\caption{\label{fig:heli}The Result of varying the Gradient Order. Top Row: Original image (left) and ground truth (right). Second row: FDOG with v=0.2 (left) and v=0.4 (right). Each row shows an increase in order in steps of 0.2 from left to right. Thus, the last row has FDOG wth v=1.8 (left) and v=2.0 (right). }
\end{figure}

Figure~\ref{fig:heli} shows the results of applying our FDOG filter with varying orders for edge detection.

%% file: learning.tex
\section{Image Segmentation using FDOG}
\label{sec:learning}


Image Segmentation is one of the most important application areas in Computer vision. Most of the image segmentation algorithms often use gradient based cues as features. In this section we apply the proposed fractional gradient based approach to an image segmentation task and evaluate its performance against state-of-the-art.

It is a common and successful practice to use Conditional Random Field (CRF) based approach in solving the segmentation task -it is also intuitively natural to think an image as a graph structure where each pixel can be considered as the nodes of the Graphical Model.
CRFs are a probabilistic framework for labeling and segmenting structured data (equation~\ref{eq:CRF}), such as sequences, trees and lattices. 

\begin{equation}
\label{eq:CRF}
P(X\mid Y)=\frac{1}{Z(Y)}\prod_C\psi(X_c,Y)\prod_i\psi(X_i,Y)
\end{equation}

The underlying idea is that of defining a conditional probability distribution over label sequences given a particular observation sequence, rather than a joint distribution over both label and observation sequences. 

However like most of the interesting problems  exact inference is usually intractable or NP- hard problem due to the high tree-width of the graph structure. Learning the graph structure is also intractable in most cases~\cite{Koller:2009:PGM:1795555}. Many previous research work has been focused on approximating the likelihood. We particularly focus on the recent work of Domke et.al.~\cite{10.1109/TPAMI.2013.31} where the Learning/ parameter fitting is viewed as repeated inference problem using the marginalization based loss function, which directly relates to the quality of prediction of a given marginal inference algorithm~\cite{10.1109/TPAMI.2013.31} also it seems to work better intuitively because it takes into account the approximation error and also it is robust to model miss-specification.  


After learning a CRF model $P(Y\mid X)$ the query is to find the most likely configuration $\hat{x}$ given the observations. One option is to use the idea of Bayes estimator using the notion of utility function ~\cite{nikolova2007model} which specifies the happiness of predicting $\hat{x}$ if $X^*$ was the true output.Then the problem reduces to an optimization task where one finds the $\hat{X}$ that maximizes the Utility function. 
\begin{equation}
\hat{X}=\arg \!\max_X\sum_{x^{*}}P(X^*\mid Y)U(X,X^*)
\end{equation}
It is also shown that using the utility function as the indicator function that is 1 when output equals $X^*$ and zero otherwise, then it reduces to MAP estimate ~\cite{nikolova2007model} i.e. if, $U(X,X^*)=I[X=X^*]$
\begin{equation}
\hat{x}=\arg\!\max_X P(X\mid Y) 
\end{equation}
However, in real examples it might require humongous amount of data for accurate prediction. Thus, in practice, a utility function that maximizes the Hamming distance i.e. the number of components in the output that are correct, is used. This framework is popularly known as Maximum Posterior Marginal(MPM)[48]. 
In the context of the current problem of learning the graphical model parameters, we view Learning from an Empirical risk Minimization framework as proposed in \cite{10.1109/TPAMI.2013.31}. where as usual the Risk is given by equation~\ref{eq:risk}.

\begin{equation}
\label{eq:risk}
R( \theta )=\sum_{\hat{x}}L(\theta,\hat{x})
\end{equation}
where the Loss function $L(\theta,\hat{x})$ explains how good the fit is with parameter $\theta$.



One can find exact marginals using tree structured graphs- which is not practical in case of graphs with high tree width such as the current problem in consideration because in that case finding the exact likelihood and its gradient becomes computationaly hard due to the log partition function and the marginals. One widely used approach to tackle this problem has been the use of markov chain Monte carlo to approximate the marginals or using Constructive Divergence based approaches\cite{carreira2005contrastive}\cite{stewart2008learning} which also suffers from large complexity
A more recent approach is to use approximate inference methods which can be viewed as approximating the partition function in the likelihood itself. Then it can be shown that the approximate marginals itself become the exact gradient of the Surrogate Loss. In recent literature Surrogate Loss seems to be widely used where the marginals being approximated by either Mean Field, Loopy Belief Propagation(LBP) or Tree-reweighted Belief Propagation (TRW).\cite{wainwright2006estimating}\cite{weinman2008efficiently}
If approximate log partition function is used that bounds the true log partition function then Surrogate Likelihood is proved to be bounding the true likelihood where Mean field based Surrogate Likelihood Upper bounds and TRW based Surrogate Likelihood lower bounds. Other than these there is Expectation maximization(EM) based approaches which is appropriate in case of incomplete data\cite{russell2006using}, saddle point based approximations\cite{tighe2010superparsing}, pseudo-likelihood based approaches\cite{kim2007robust} and Piecewise Likelihood\cite{koppula2011semantic}  etc. Another issue to be taken into consideration is model mis-specification which often might be unavoidable due to the complex nature of problems or deliberately in case when the true model has many parameters to fit the data which leads to many degrees of freedom thus reduction becomes necessary.

For this particular problem we chose to train the model using a Clique Loss as in the present setting it is intuitively more straightforward to think about clique based loss rather than univariate loss as proved by Wainwright and Jordan et.al.\cite{wainwright2008graphical} that due to the standard moment matching criterion of exponential families if the clique marginals are correct the joint distribution has to be correct though the joint distribution might be far off perfect even if the univariate marginals are perfect.It is defined mathematically as-
\begin{equation}
L(\theta,x)=-\sum_Clog\mu(x_c;\theta)
\end{equation}
which can be viewed minimizing the empirical risk of the mean KL-divergence of the true clique marginals to
the predicted ones.
The next step is to do the inference which turns out to be tricky for graphs with high tree width which requires two operations- firstly the evaluation of the loss function, which is pretty straight forward and can be done simply by plugging the marginal obtained by running the inference algorithm into the loss function and secondly, one requires to know the gradient $\frac{dL}{d\theta}$. The gradient can be computed by solving a set of sparse linear equation which assumes that the optimization problem is exactly solved ,though in practice one has to set a threshold to truncate which can be done very elegantly with a much lower computational expense by forming the learning
objective in terms of the approximate marginals obtained
after a fixed number of iterations. ~\cite{10.1109/TPAMI.2013.31}

It leads to a series of simple structured steps viz. Inputting parameters, applying the iterations of either
TRW or mean field, computing predicted marginals,
and finally the loss are all differentiable operations.
Here we use back-propagating Tree Re-weighted Belief Propagation Framework which calculates the marginals and plug them into the marginal based loss after a certain number of iterations even if the inference algorithm did not converge, which is possible because each step is differentiable.  

After execution of Back TRW 
the parameter $\theta(x_c)$ for the clique based loss can be represented mathematically as  
\begin{equation}
\overleftarrow{\theta}(x_c)=\frac{dL}{d\theta(x_c)}
\end{equation}

%% file: results.tex
\subsection{Experimental Results of using FDOG for segmentation}
\label{sec:results}

In order to experiment with our proposed FDOG based method we have chosen the Stanford background dataset which has 715 images of outdoor scenes taken from different publicly available datasets like LabelMe, MSRC, PASCAL VOC and Geometric Context; each image of resolution approximately 240 * 320. Most pixels are labeled one of the eight classes with some unlabeled. 
All the experiments have been done on a Dual Core, 3GHz Processor with parallel tasks running on both the processor.
We have sticked with the simple features used in the state-of-the-art algorithm \cite{10.1109/TPAMI.2013.31} with replacing the gradient computations by our method in order to demonstrate the success of the proposed algorithm. 
The experimental method has the following flow chart.
\begin{enumerate}
\item We have used a pairwise 4 connected grid model.
\item We found the appropriate order of the Gradient Operator to be 0.7 using the score Function described in Section~\ref{sec:imsegment}. To run over all images and compute the approximate one decimal approximate of FDOG it took about 5 minutes.
\item For the unary features we first computed RGB intensities of each pixel, along with the normalized 
vertical and horizontal positions. We expand these initial features into a larger set using sinusoidal expansion, specifically use sin(c.s) and cos(c.s) where s is the initial feature mentioned for all binary vectors c of appropriate lengths leading to total 64 features. \cite{konidaris2011value}
\item Compute FHOG (fractional HOG)i.e. Histogram of Oriented gradient Features but the gradients are found using the proposed method i.e. using a third order approximate kernel using ~\ref{eq:eq1} where the order was taken as 0.6.
\item For Edge Features between pixels i and j we considered 21 base features comprsinga constant of one,Euclidian Norm of RGB value difference dicretized into 10 levels and maximum response of FDOG at i or j over three channels and aging binning into 10 dicrete levels and one feature based on the difference of RGB intensities.
\item All methods are trained using TRW to fit the Clique Logistic Loss. 
\end{enumerate}

Apart from using FDOG we also change the Optimization algorithm to R-PROP for improved computational efficiency.

\subsection{Performance}
\begin{figure}
\centering
\includegraphics[scale=0.2]{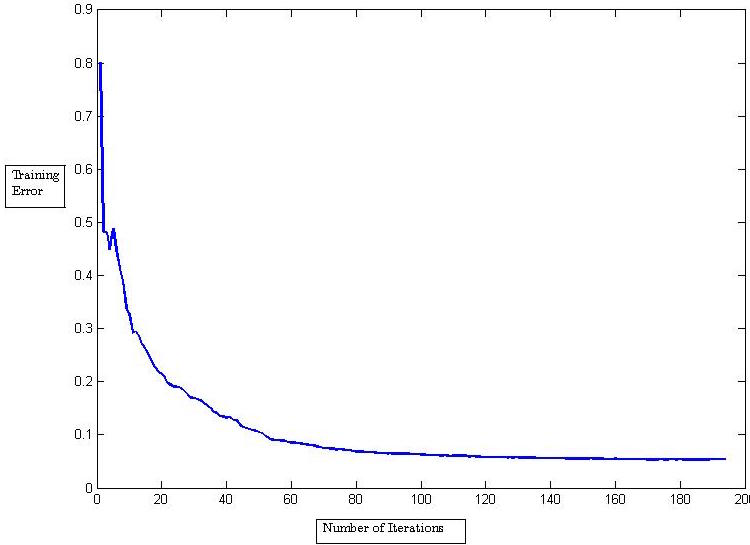}
\caption{\label{fig:tperf}Training Performance of the Model.}
\end{figure}

The model is trained with 90 Images in a Monte Carlo Manner with 10 fold cross validation splitting the Training Set into Training and Validation in 90:10 ratio and the data was also shuffled randomly to avoid pathological ordering. The training took around 5 hours on a Intel dual core 3GHz Processor. The training performance is shown in Figure~\ref{fig:tperf}

\begin{figure}
\centering
\includegraphics[scale=0.45]{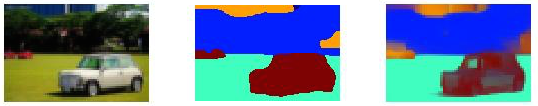}
\includegraphics[scale=0.45]{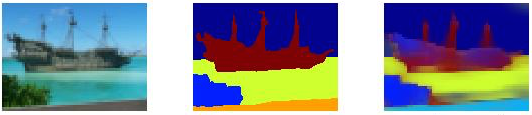}
\includegraphics[scale=0.45]{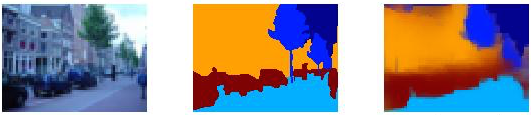}
\includegraphics[scale=0.45]{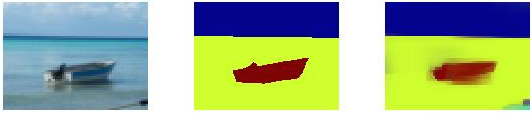}
\includegraphics[scale=0.45]{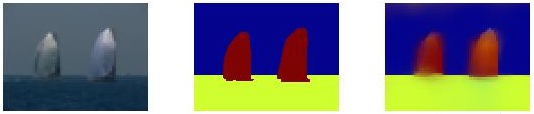}
\includegraphics[scale=0.45]{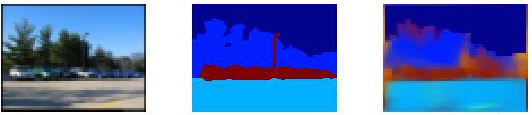}
\includegraphics[scale=0.45]{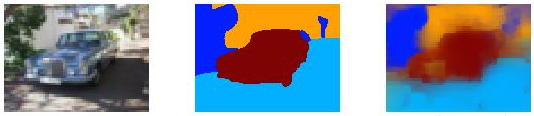}
\includegraphics[scale=0.364]{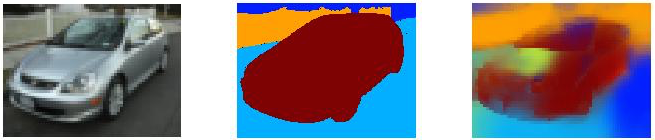}
\includegraphics[scale=0.364]{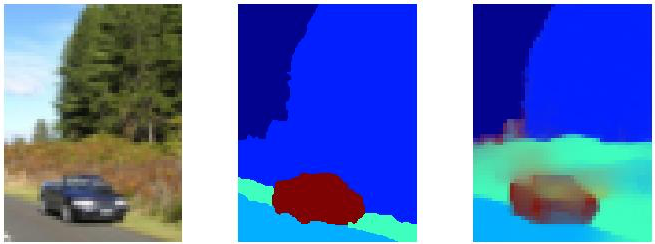}
\includegraphics[scale=0.364]{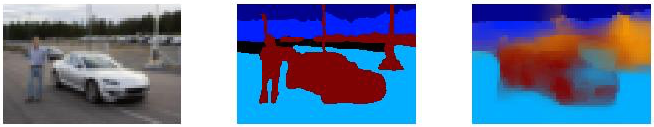}
\caption{\label{fig:results}Image segmentation. Each row, from left to right: original image, true label and estimated label.}
\end{figure}

The TRW BP approach by Domke et.al ~\cite{10.1109/TPAMI.2013.31} using same set of features but HOG and and edge detectors use DOG is the state of the art 22.1\% average error on test data. We bettered that in our experiments reducing the error by almost 10\%. Our average error on the test data turned out to be 20.8\%. Table~\ref{table:table2} shows the result on Stanford Dataset. Some visual results on test data are shown in Figure~\ref{fig:results}

\begin{table}

  \caption{Comparison with different algorithms}
  \centering
\begin{tabular}{|l|c|r|}
\hline
Algorithm & Avg. Accuracy\\\hline
Region-based energy\cite{konidaris2011value} & 65.5 \\\hline
Stacked Labeling \cite{gould2009decomposing} & 66.2\\\hline
RGB-D \cite{munoz2010stacked} & 74.5\\\hline
TRW BP+Clique Loss \cite{10.1109/TPAMI.2013.31} & 77.9\\\hline 
TRW BP+ CLique Loss+ FDOG [us]& 79.2\\\hline \hline
\end{tabular}
\label{table:table2}
\end{table}

\section{Discussion and Future Work}
\label{sec:conclusion}
As can be seen from the demonstrated results we have done quite well on the difficult Stanford Backgrounds Dataset. Though 20.8\% error is quite high, there is a lot of room for improvement. The reason for this improvement is the efficient calculation of Gradient Features and it can improve the performance of any of the existing algorithms that are based on Gradient Features. Some of the future work in this might be to extend the concept directly from Frequency Domain and see how it affects the performance. Some, other datasets can be explored. Also, one important option is to apply the Texture Based Approach using this framework as intuitively it seems to be a more realistic idea in case of Contour Based Edge Detection. We believe that using this unified framework the Texton Based approach can also be improved by a few notch.